\documentclass[conference]{IEEEtran}
\IEEEoverridecommandlockouts
\usepackage{cite}
\usepackage{amsmath,amssymb,amsfonts}
\usepackage{algorithmic}
\usepackage{graphicx}
\usepackage{textcomp}
\usepackage{xcolor}
\usepackage{booktabs}       
\usepackage{comment} 
\usepackage{multirow}
\usepackage{rotating}
\usepackage[caption=false,font=scriptsize,labelfont=sf,textfont=sf]{subfig}
\usepackage{url}            
\def\BibTeX{{\rm B\kern-.05em{\sc i\kern-.025em b}\kern-.08em
    T\kern-.1667em\lower.7ex\hbox{E}\kern-.125emX}}
\begin{document}

\title{Loss Surface Modality of Feed-Forward Neural Network Architectures
}
\author{\IEEEauthorblockN{Anna Sergeevna Bosman}
\IEEEauthorblockA{\textit{Department of Computer Science} \\
\textit{University of Pretoria}\\
Pretoria, South Africa \\
https://orcid.org/0000-0003-3546-1467\\
annar@cs.up.ac.za}
\and
\IEEEauthorblockN{Andries Petrus Engelbrecht}
\IEEEauthorblockA{\textit{Department of Industrial Engineering} \\
	\textit{Computer Science Division} \\
\textit{Stellenbosch University}\\
Stellenbosch, South Africa \\
engel@sun.ac.za}
\and
\IEEEauthorblockN{Mard\'e Helbig}
\IEEEauthorblockA{\textit{School of Information and} \\
\textit{Communication Technology} \\
\textit{Griffith University}\\
Southport, Australia \\
m.helbig@griffith.edu.au}
}

\maketitle

\begin{abstract}
It has been argued in the past that high-dimensional neural networks do not exhibit local minima capable of trapping an optimisation algorithm. However, the relationship between loss surface modality and the neural architecture parameters, such as the number of hidden neurons per layer and the number of hidden layers, remains poorly understood. This study employs fitness landscape analysis to study the modality of neural network loss surfaces under various feed-forward architecture settings. An increase in the problem dimensionality is shown to yield a more searchable and more exploitable loss surface. An increase in the hidden layer width is shown to effectively reduce the number of local minima, and simplify the shape of the global attractor. An increase in the architecture depth is shown to sharpen the global attractor, thus making it more exploitable.
\end{abstract}


\section{Introduction}
Neural network (NN) performance is known to depend on the chosen NN architecture, i.e., the number of neurons, hidden layers, and the structure of connections~\cite{ref:Benardos:2007,ref:Huang:2003,ref:Huang:2006,ref:Lapedes:1988,ref:Wilamowski:2009}. A NN with too few trainable parameters will not be able to fit complex non-linear data, and a NN with an excessive number of trainable parameters was argued to be prone to overfitting~\cite{ref:Wilamowski:2009}. However, recent advances in deep learning challenged our understanding of the relationship between the neural architecture and the performance of a NN, empirically and theoretically showing that excessive complexity often leads to superior results~\cite{ref:Caruana:2001,ref:Nguyen:2017, ref:Sagun:2015}. Sagun et al.~\cite{ref:Sagun:2015} empirically observed that no high error local minima were detected when the NN architecture was over-parametrised. Theoretical studies have also been published showing that over-parametrised models do not exhibit high error local minima~\cite{ref:Ballard:2017,ref:Haihao:2017}. In fact, recent studies claim that using an excessively large hidden layer (larger than the number of training points in the dataset) guarantees that almost all local minima will be globally optimal~\cite{ref:Nguyen:2017}. The opposite is also true: deep and ``skinny'' NNs, i.e., NNs with a limited number of hidden neurons per layer, were theoretically shown to not exhibit the universal approximator properties~\cite{ref:Johnson:2018}.

With the number of neural architecture search algorithms on the rise~\cite{elsken2019neural}, it becomes increasingly important to improve our understanding of the NN loss surfaces~\cite{ref:Choromanska:2015a, ref:Kordos:2004, ref:Shen:2016}, and the implications of width (i.e. number of hidden neurons per layer) and depth (i.e. number of layers) on the resulting optimisation problem.

This study aims to investigate NN loss surfaces under various NN architecture settings using a sampling-based technique developed for fitness landscape analysis (FLA). Stationary points are discovered and classified into minima, maxima, and saddles using Hessian matrix analysis. A simple visualisation method proposed in~\cite{ref:Bosman:2019} is used, which enables intuitive insights into the nature of the loss surfaces. The experiments correlate well with the current theoretical insights, and provide ground for further hypothesis. 

The main contributions of this paper are as follows:
\begin{itemize}
	\item Empirical evidence for the existing theoretical conjectures~\cite{ref:Johnson:2018,ref:Nguyen:2017} is obtained.
	\item An increase in the problem dimensionality is shown to yield a more searchable and more exploitable loss surface. 
	\item An increase in the architecture width is shown to effectively reduce the number of local minima, and simplify the shape of the global attractor. 
	\item An increase in the architecture depth is shown to sharpen the global attractor, thus making it more exploitable.
\end{itemize}

The rest of the paper is structured as follows: Section~\ref{sec:arch:fla} briefly discusses fitness landscape analysis in the context of NNs. Section~\ref{sec:arch:setup} describes the experimental procedure of this study. Section~\ref{sec:arch:basins} presents a visual analysis of stationary points and basins of attraction associated with the various NN architectures. Section~\ref{sec:arch:conclusions} concludes the paper.

\section{Fitness landscape analysis of neural networks}\label{sec:arch:fla}
Fitness landscape analysis (FLA) was first applied in the evolutionary context, where quantitative metrics were proposed to study the optimisation landscapes of combinatorial problems~\cite{ref:Jones:1995,ref:Merz:2000}. FLA techniques were soon extended to continuous search spaces~\cite{ref:Malan:2009,ref:Malan:2014,ref:Munoz:2015,ref:Sun:2014}. FLA techniques aim to estimate various fitness landscape properties, such as ruggedness, neutrality, modality, and searchability, based on multiple samples of the search space. An objective function value is calculated for each sampled point, and the relationship between the spatial and the qualitative characteristics of the sampled points is analysed. Properties of a fitness landscape captured by the samples are considered to be an approximation of the global fitness landscape properties. 

The NN search space is defined as all possible real-valued weight combinations. Thus, weight space can be sampled to make conclusions about the search space properties. This study considers the NN loss surface modality, i.e., local and global minima properties. The FLA technique called loss-gradient (l-g) clouds~\cite{ref:Bosman:2019} is used to visualise the basins of attraction of NN loss surfaces in a 2-dimensional projection. 

Sampling is performed using progressive gradient walks~\cite{ref:Bosman:2018}. A progressive gradient walk uses the numeric gradient of the loss function to determine the direction of each step. The size of the step is randomised per dimension within predefined bounds. The progressive gradient walk algorithm is summarised as follows:
\begin{enumerate}
	\item Gradient vector $\vec{g}_l$ is calculated for a point $\vec{x}_l$.
	\item A binary direction mask $\vec{b}_l$ is extracted from
	$\vec{g}_l$ as follows:
	\begin{equation*}
	b_{lj} =\begin{cases}
	0 & \text{if $g_{lj}<0$},\\
	1 & \text{otherwise},
	\end{cases}
	\end{equation*}
	where $j \in \{1,\dots,m\}$ for the $m$-dimensional vector $\vec{g}_l$.
	\item The progressive random walk algorithm, proposed in~\cite{ref:Malan:2014b}, is used to generate the next step $\vec{x}_{l+1}$. A single step of a progressive random walk can be defined as randomly generating an $m$-dimensional step vector $\Delta\vec{x}_l$, such that $\Delta{x}_{lj}\in [0,\varepsilon]$~$\forall j \in
	\{1,\dots,m\}$, and setting the sign of each $\Delta{x}_{lj}$ according to the corresponding ${b}_{lj}$:
	\begin{equation*}
	\Delta{x}_{lj} :=\begin{cases}
	-\Delta{x}_{lj} & \text{if ${b}_{lj} = 0$},\\
	\Delta{x}_{lj} & \text{otherwise}.
	\end{cases}
	\end{equation*} 
	To generate the next step, $\vec{x}_{l+1}$, the current step $\vec{x}_{l}$ is modified by adding $\Delta\vec{x}_l$:
	\begin{equation*}\label{eq:simple}
	\vec{x}_{l+1} = \vec{x}_{l} + \Delta\vec{x}_l
	\end{equation*}
\end{enumerate}
The progressive gradient walk algorithm requires one parameter to be
set: the maximum dimension-wise step size, $\varepsilon$.

The advantage of this sampling approach compared to random sampling or NN training algorithm sampling is that gradient information is combined with stochasticity, preventing convergence, yet guiding the walk towards areas of higher fitness (i.e. lower error). 

Once the sampling is complete, a 2-dimensional scatterplot is generated, with the sampled loss values on the $x$-axis, and the corresponding gradient vector magnitude values on the $y$-axis. All points of zero gradient are stationary points, which can be further classified into minima, maxima, or saddle points using the eigenvalues of the corresponding Hessian matrix~\cite{ref:Edwards:1973}. These scatterplots are referred to as loss-gradient (l-g) clouds for the rest of the paper. Studying the discovered stationary points in a 2-dimensional space allows for the identification of the total number of attractors, both local and global, corresponding to unique loss values. The gradient behaviour of the attractors is also visualised by the l-g clouds, and can provide useful insights into the structure of the attraction basins, such as the steepness of the basins, and the connectedness of the basins, i.e., the ability of the sampling algorithm to make a transition from a local attractor to the global attractor.

\section{Experimental set-up}\label{sec:arch:setup}
The aim of the study was to visually investigate the local minima and the associated basins of attraction exhibited by the NN architectures of varied dimensionality and structure. This section discusses the experimental set-up of the study, and is structured as follows: Section~\ref{sub:arch:bench} lists the benchmark problems used, Section~\ref{sub:arch:act} discusses the architecture settings employed, and Section~\ref{sub:arch:walks} outlines the sampling algorithm parameters, and the data recorded for each sampled point.

\subsection{Benchmark problems}\label{sub:arch:bench}
A selection of well-known classification problems of varied dimensionality were used in this study. For the sake of brevity, only two problems are discussed in this paper: the XOR problem and the MNIST problem. The XOR problem requires the NN to model the ``exclusive-or'' logical gate using four binary patterns of two inputs and one output. Despite seeming triviality, the XOR problem is not linearly separable, and thus makes a good case study for fundamental NN properties~\cite{ref:Mehta:2018}. The MNIST dataset of handwritten digits~\cite{ref:LeCun:2010} contains 70 000 examples of grey scale handwritten digits from $0$ to $9$, where 60 000 examples constitute the training set, and the remaining 10 000 constitute the test set. For the purpose of this study, the 2-dimensional input is treated as a 1-dimensional vector, with the total number of inputs equal to 784.

Note that the results obtained for other classification problems, together with the code used to run the experiments, are available at the following URL: \url{https://github.com/annabosman/fla-in-tf}

\subsection{Architectures}\label{sub:arch:act}
All experiments employed feed-forward NNs with the exponential linear unit (ELU)~\cite{ref:Clevert:2016} activation function in the hidden layers. For the binary classification problems, the sigmoid function was used in the output layer. For the multinomial classification problems, the softmax activation function was used in the output layer. Log-likelihood loss was used to calculate the NN error.

To study the influence of the hidden layer width on the NN loss surfaces, each problem was considered with $h$ minimal number of hidden neurons ($h=2$ for XOR, $h=10$ for MNIST), with twice as many hidden neurons as prescribed by the minimal architecture ($2\times h$), and with ten times as many hidden neurons ($10\times h$). These settings were chosen to simulate a minor increase in width ($2\times h$), as well as a more substantial increase corresponding to the next order of magnitude ($10\times h$).

To study the influence of the architecture depth on the NN error surfaces, 1, 2, and 3 hidden layers were considered for each hidden layer width ($h$, $2\times h$, and $10\times h$) as discussed in the preceding paragraph. The same width setting was used for each successive hidden layer.

\subsection{Sampling parameters}\label{sub:arch:walks}
Progressive gradient walks~\cite{ref:Bosman:2018} were used for the purpose of sampling. The total number of walks was set to be $2\times m$, where $m$ is the dimensionality of the search space. The walks were unbounded, but two distinct initialisation ranges were considered, namely $[-1,1]$ and $[-10,10]$. Two granularity settings were used throughout the experiments: micro, where the maximum step size, $\varepsilon$, was set to 1\% of the initialisation range, and macro, where $\varepsilon$ was set to 10\% of the initialisation range. Micro walks performed 1000 steps each, and macro walks performed 100 steps each. To calculate the training ($E_t$) and the generalisation ($E_g$) errors for MNIST, random batches of 100 patterns were randomly sampled from the respective training and test sets. The same data subsets were used to calculate the average classification accuracy obtained at the last step of the gradient walks. All experiments were run on a single node of a computing cluster with 24 Intel 5th generation CPUs and 128 GB RAM.

\section{Empirical results}\label{sec:arch:basins}
This section presents the analysis of sampled local minima and the corresponding basins of attraction as captured by the progressive gradient walks for the various NN architectures. L-g clouds, discussed in Section~\ref{sec:arch:fla}, are employed for the purpose of this study. Sections~\ref{sec:arch:xor} and~\ref{sec:arch:mnist} present the analysis of the various NN architectures for XOR and MNIST, respectively.

\subsection{XOR}\label{sec:arch:xor}
\begin{figure}[htb]
	\begin{center} \includegraphics[width=0.4\textwidth]{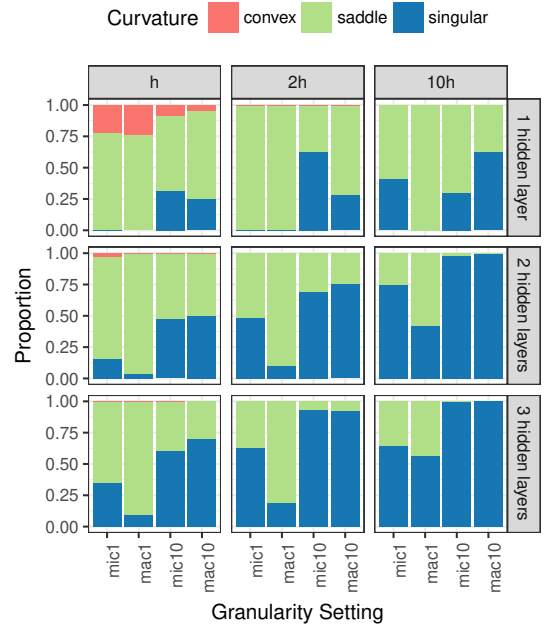}
		\caption{Histogram representation of the curvature information sampled by the progressive gradient walks for the XOR problem for various NN architecture settings. Mic and mac correspond to micro and macro granularities, 1 and 10 correspond to $[-1,1]$ and $[-10,10]$ initialisation ranges.}\label{fig:xor:arch:curv}
	\end{center}
\end{figure}

Curvature information was obtained by calculating the eigenvalues of the Hessian for each sampled point. Figure~\ref{fig:xor:arch:curv} summarises the curvature information obtained for the different architectures and sampling settings considered. Each bar in the plot corresponds to a distinct granularity setting, and is colourised proportionally to the curvature of the sampled points. The plots are grouped horizontally according to the layer width ($h, 2\times h$, and $10\times h$), and vertically according to the network depth (1, 2, and 3 hidden layers). 
According to Figure~\ref{fig:xor:arch:curv}, an increase in width caused a reduction in convexity and an increase in flatness (singular Hessians). Indeed, the addition of extra neurons to a minimal architecture introduces unnecessary, or redundant weights. Since more compact solutions, i.e., solutions with fewer weights, are embedded in over-parametrised architectures~\cite{ref:Mehta:2018}, the discovery of such solutions will cause the unnecessary neurons to be disabled, thus introducing flatness.

Figure~\ref{fig:xor:arch:curv} shows that an increase in depth had a similar effect: Convexity decreased, and flatness increased. In fact, according to Figure~\ref{fig:xor:arch:curv}, an increase in depth increased the flatness more rapidly than an increase in width. The rapid increase in flatness associated with deeper architectures is attributed to the inter-dependent variable structure of feed-forward NNs. Since each layer propagates the signals to the next layer, each layer has the ability to set the incoming signals to zero. In other words, if a neuron in a later layer saturates, the effects of saturation will influence the contribution of the weights in all preceding layers. 

An increase in flatness due to an increase in dimensionality, whether by adding extra neurons, or by adding extra layers, agrees with the findings of Sagun et al.~\cite{ref:Sagun:2017}, where Hessian analysis of over-parametrised NNs was performed for the first time. A question that remains to be answered is whether the modality of the NN loss surface changes when hidden neurons/hidden layers are added, and whether the effect of increased width differs from the effect of increased depth. 

\begin{figure*}[!tb]
	\begin{center}
		\begin{subfloat}[{$h = 2$,  1 hidden layer }\label{fig:xor:b1:h:l1:micro}]{    \includegraphics[width=0.45\textwidth]{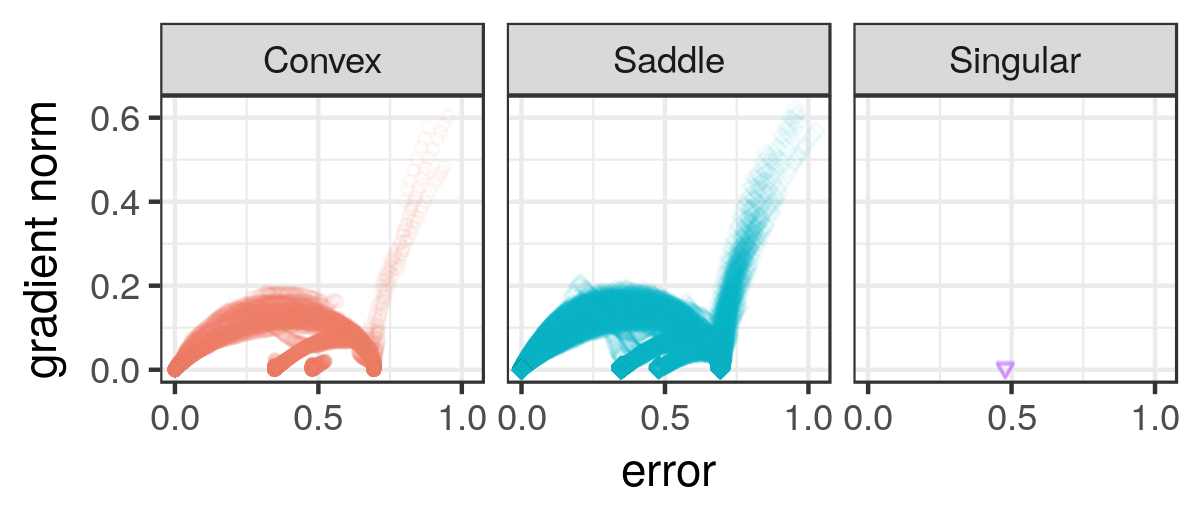}}
		\end{subfloat}\hfill
		\begin{subfloat}[{$h = 2$, 2 hidden layers }\label{fig:xor:b1:h:l2:micro}]{    \includegraphics[width=0.45\textwidth]{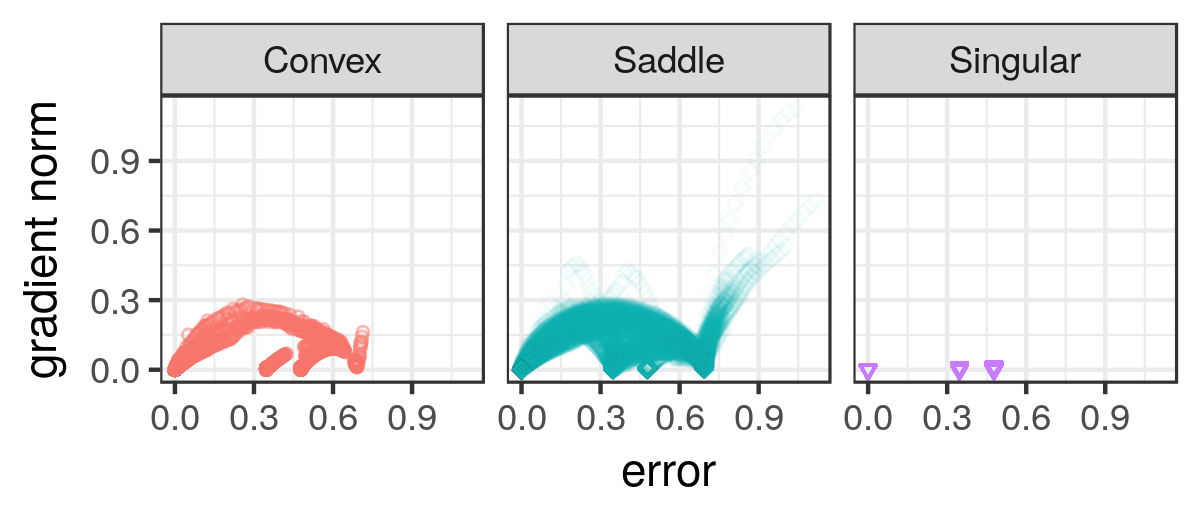}}
		\end{subfloat}\\
		\begin{subfloat}[{$2\times h = 4$, 1 hidden layer }\label{fig:xor:b1:2h:l1:micro}]{    \includegraphics[width=0.45\textwidth]{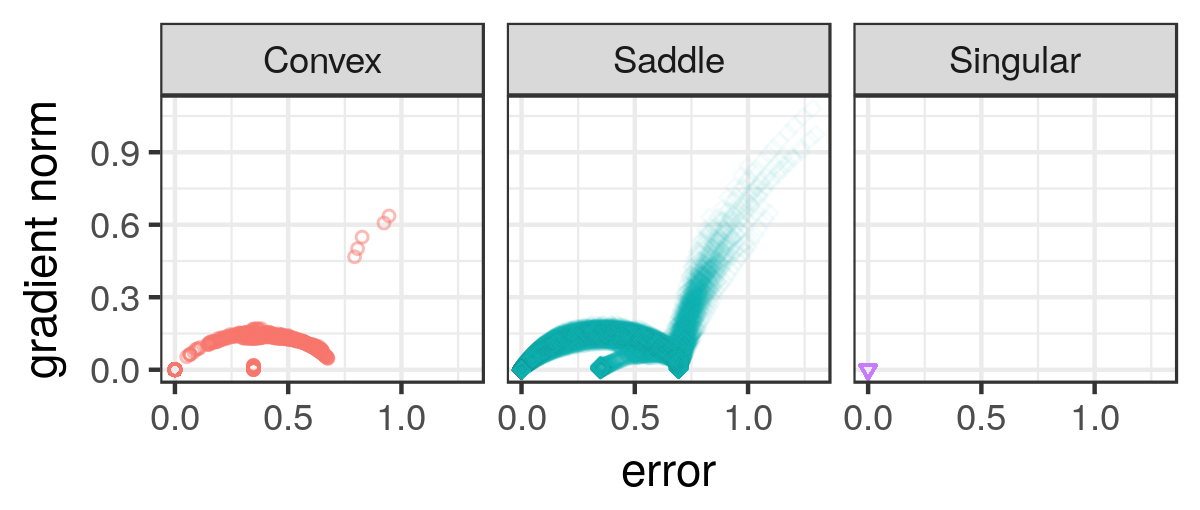}}
		\end{subfloat}\hfill
		\begin{subfloat}[{$2\times h = 4$, 2 hidden layers }\label{fig:xor:b1:2h:l2:micro}]{    \includegraphics[width=0.32\textwidth]{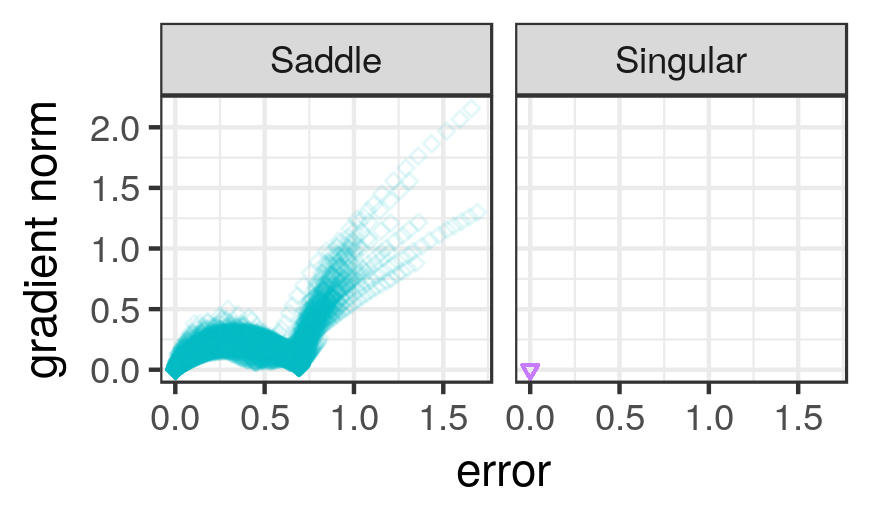}}
		\end{subfloat}\\
		\begin{subfloat}[{$10\times h = 20$,  1 hidden layer }\label{fig:xor:b1:10h:l1:micro}]{    \includegraphics[width=0.32\textwidth]{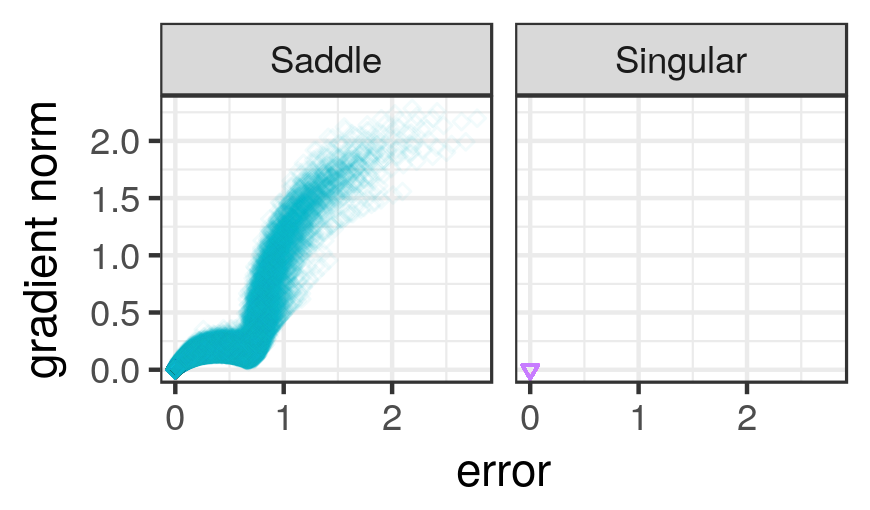}}
		\end{subfloat}\hfill
		\begin{subfloat}[{$10\times h = 20$, 2 hidden layers }\label{fig:xor:b1:10h:l2:micro}]{    \includegraphics[width=0.32\textwidth]{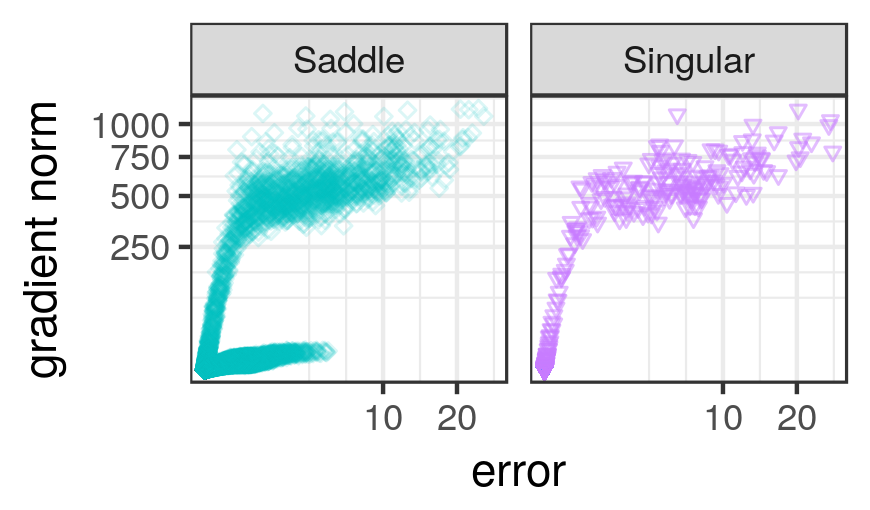}}
		\end{subfloat}
		\caption{Loss-gradient clouds for the micro gradient walks ($\varepsilon = 0.02$) initialised in the $[-1,1]$ range for the XOR problem for the various hidden layer widths.}\label{fig:xor:b1:micro:arch:l1}
	\end{center}
\end{figure*}
\subsubsection{The effect of width}

Figure~\ref{fig:xor:b1:micro:arch:l1} shows the l-g clouds obtained for 1- and 2-hidden layer architectures of varied width. Figure~\ref{fig:xor:b1:micro:arch:l1} shows that the number of convex attractors, i.e., local minima (represented by points of zero gradient norm and non-zero error) reduced as more hidden neurons were added to a layer. For $h=2$, four attractors, three of them constituting local minima, were observed (Figures~\ref{fig:xor:b1:h:l1:micro} and~\ref{fig:xor:b1:h:l2:micro}). For $h=4$, only three attractors were detected for the single hidden layer architecture. Two of the attractors constituted local minima (Figure~\ref{fig:xor:b1:2h:l1:micro}). Finally, for $h=20$, two non-convex attractors were observed (Figure~\ref{fig:xor:b1:10h:l1:micro}), thus local minima were eliminated altogether. 

The same trend is evident for the 2-hidden layer architectures: The number of attractors decreased as more hidden neurons were added. Figure~\ref{fig:xor:b1:10h:l2:micro} also indicates that excessive width yielded a split into two clusters, namely the high gradient cluster associated with flat curvature in individual dimensions, and a low gradient cluster of saddle curvature. Lack of curvature in some dimensions means that those particular dimensions did not contribute to the final loss value. In the case of NNs, each dimension corresponds to a weight. A non-contributing weight indicates that the signal generated by that weight is zeroed somewhere in the architecture, and any change in the weight value would not have an effect on the NN output. In a $2\times 20 \times 20 \times 1$ feed-forward architecture used for the XOR problem, a large number of neurons can be safely disabled without damaging the quality of the model, since the optimal architecture for XOR is $2\times 2 \times 1$. Therefore, these unnecessary neurons contribute to the flatness of the resulting NN loss surface.  It has been observed in the past that optima for smaller NN architectures are embedded in the weight space of larger NN architectures~\cite{ref:Mehta:2018}. Thus, flat areas around the global minimum are attributed to the non-contributing weights, which are associated with implicit regularisation.

Thus, an increase in the hidden layer width yielded a decrease in the number of convex attractors for the XOR problem, and reduced the number of local minima attractors. This observation correlates with the recent theoretical study of Nguyen and Hein~\cite{ref:Nguyen:2017}, where using more hidden neurons than the number of training points was shown to guarantee that most local minima would be globally optimal.

\subsubsection{The effect of depth}
\begin{figure*}[!tb]
	\begin{center}
		\begin{subfloat}[{1 hidden layer, $h = 2$  }\label{fig:xor:b1:h:l1:macro}]{    \includegraphics[width=0.32\textwidth]{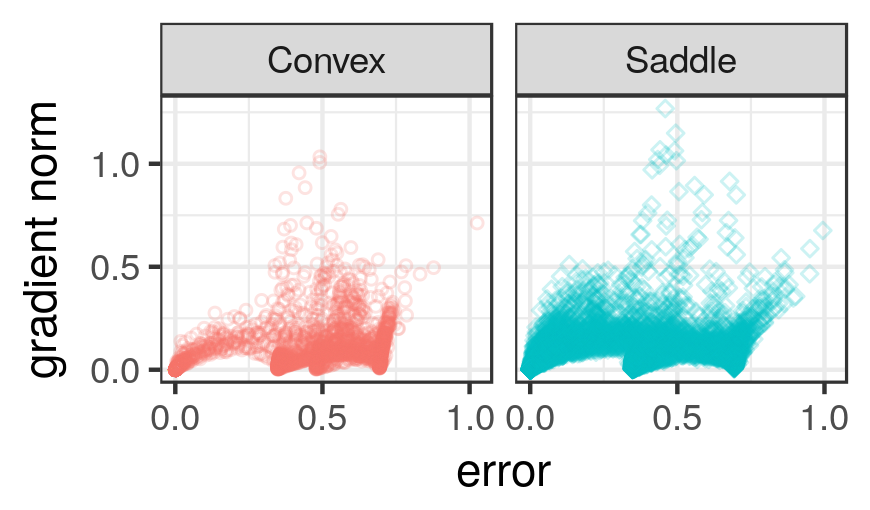}}
		\end{subfloat}\hfill
		\begin{subfloat}[{1 hidden layer, $2\times h = 4$  }]{    \includegraphics[width=0.45\textwidth]{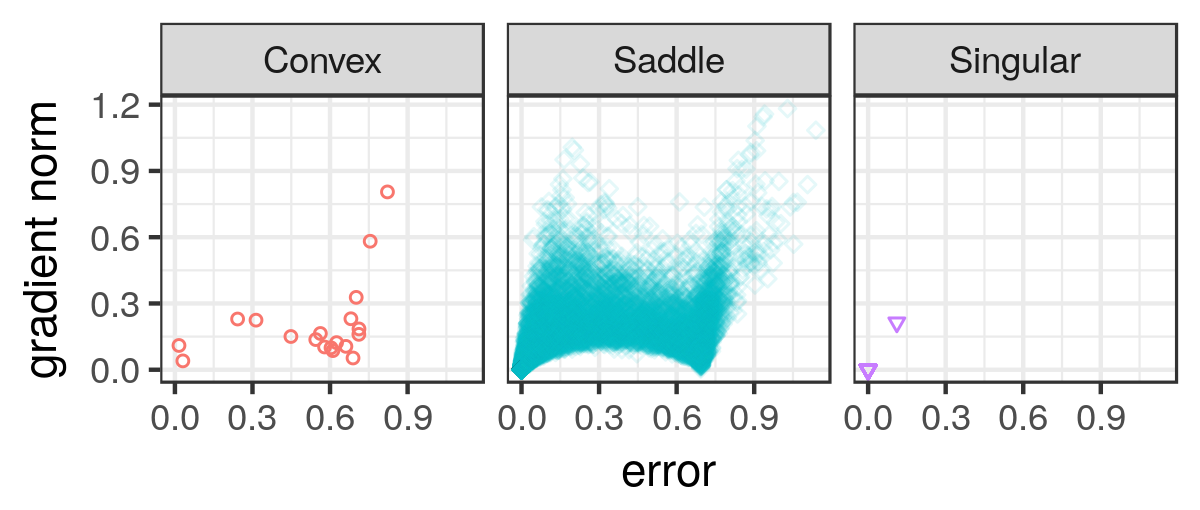}}
		\end{subfloat}\\
		\begin{subfloat}[{2 hidden layers, $h = 2$  }\label{fig:xor:b1:h:l2:macro}]{    \includegraphics[width=0.45\textwidth]{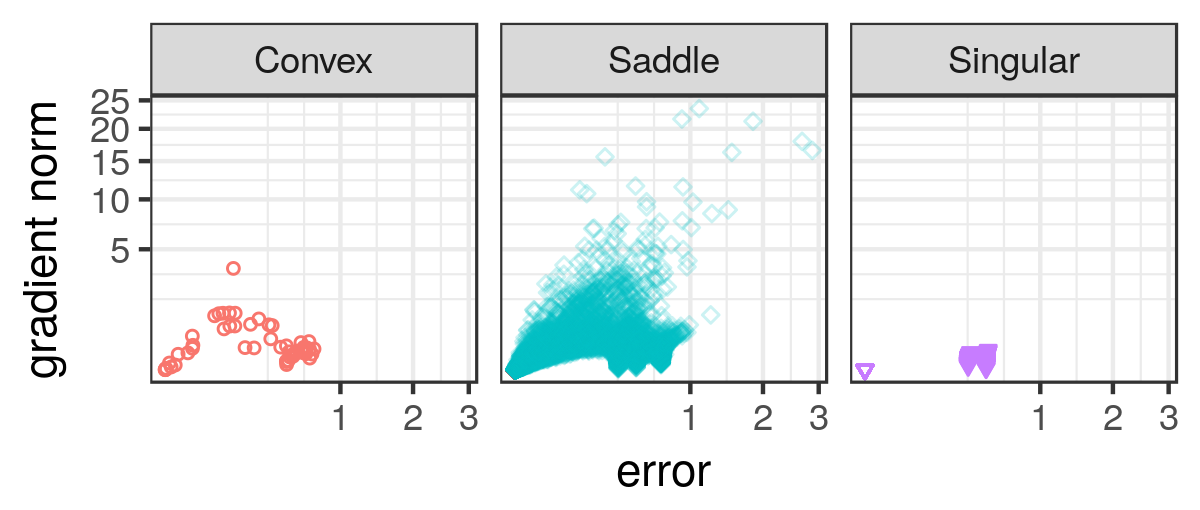}}
		\end{subfloat}\hfill
		\begin{subfloat}[{2 hidden layers, $2\times h = 4$ }]{    \includegraphics[width=0.32\textwidth]{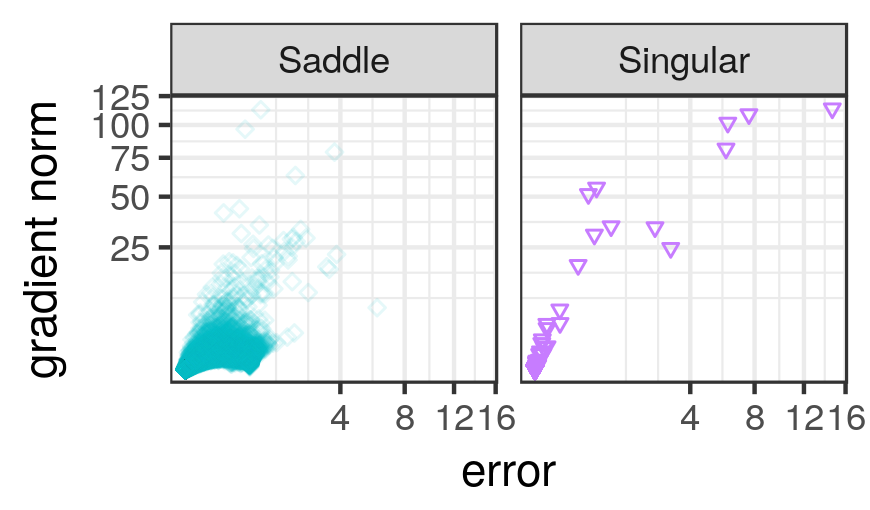}}
		\end{subfloat}\\
		\begin{subfloat}[{3 hidden layers, $h = 2$  }\label{fig:xor:b1:h:l3:macro}]{    \includegraphics[width=0.45\textwidth]{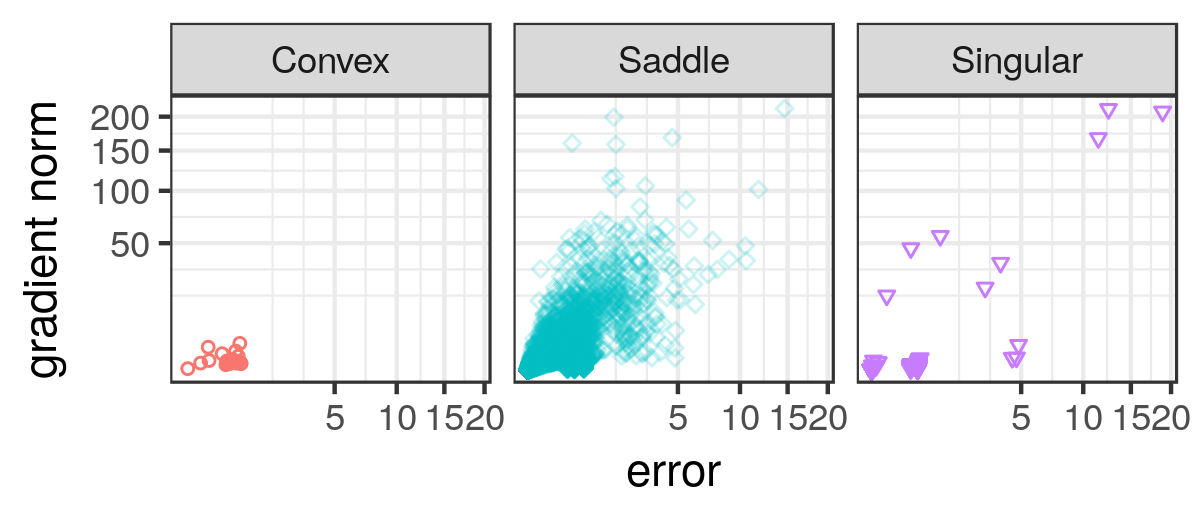}}
		\end{subfloat}\hfill
		\begin{subfloat}[{3 hidden layers, $2\times h = 4$ }]{    \includegraphics[width=0.32\textwidth]{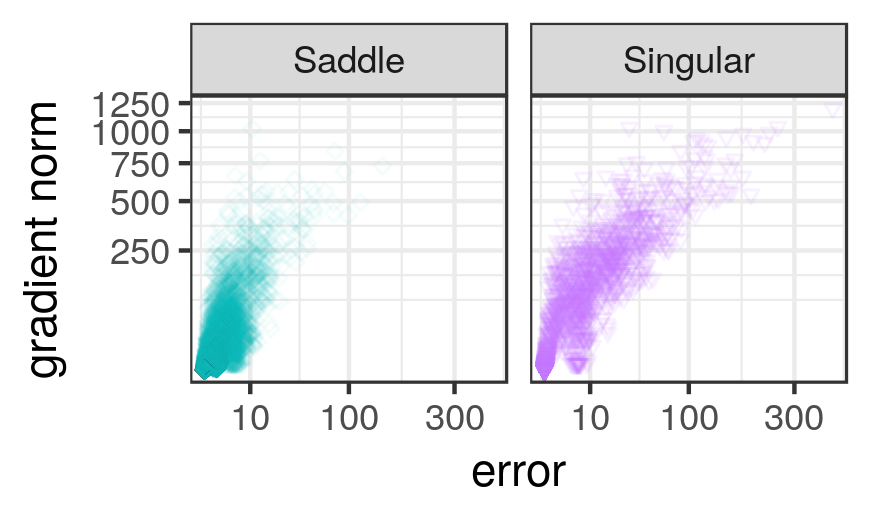}}
		\end{subfloat}
		\caption{Loss-gradient clouds for the macro gradient walks ($\varepsilon = 0.2$) initialised in the $[-1,1]$ range for the XOR problem for the various architecture depths.}\label{fig:xor:b1:macro:arch:h2}
	\end{center}
\end{figure*}

Figure~\ref{fig:xor:b1:macro:arch:h2} shows the l-g clouds obtained for the $[-1,1]$ macro walks executed on the loss surface yielded by a NN architecture of varied depth (1, 2, and 3 hidden layers). Figure~\ref{fig:xor:b1:h:l1:macro} shows that four stationary convex attractors were discovered for the single hidden layer architecture with $h=2$. Three of the attractors constituted local minima. The addition of the second hidden layer (Figure~\ref{fig:xor:b1:h:l2:macro}) decreased the convexity and increased the flatness around the attractors, but the total number of attractors remained the same. The addition of the third hidden layer (Figure~\ref{fig:xor:b1:h:l3:macro}) also yielded exactly four zero-gradient attractors. Thus, the number of hidden layers did not change the modality of the landscape for the XOR problem, i.e., the number of attractors of a unique error value remained the same. Figure~\ref{fig:xor:b1:macro:arch:h2} shows that the same behaviour was observed for the architectures with four hidden neurons per layer: For the varied NN depth, exactly two attractors were observed.

Even though the number of attractors was not altered by the NN depth, the properties of the said attractors were affected. In addition to the decreased convexity and increased flatness, a drastic increase in the gradient magnitudes was observed for deeper architectures. Figure~\ref{fig:xor:b1:macro:arch:h2} shows that each new layer increased the maximum gradient by an order of magnitude. Figure~\ref{fig:xor:b1:micro:arch:l1} shows that the increase in width also caused an increase in the gradient magnitudes, but not as drastic, especially for the single hidden layer architecture. The range of the error values also increased rapidly for each new layer added.

Thus, an increase in width decreased the number of observed minima for the XOR problem, and an increase in depth did not alter the modality properties of the loss surface.

\subsection{MNIST}\label{sec:arch:mnist}
\begin{figure}[!htb]
	\includegraphics[width=0.49\textwidth]{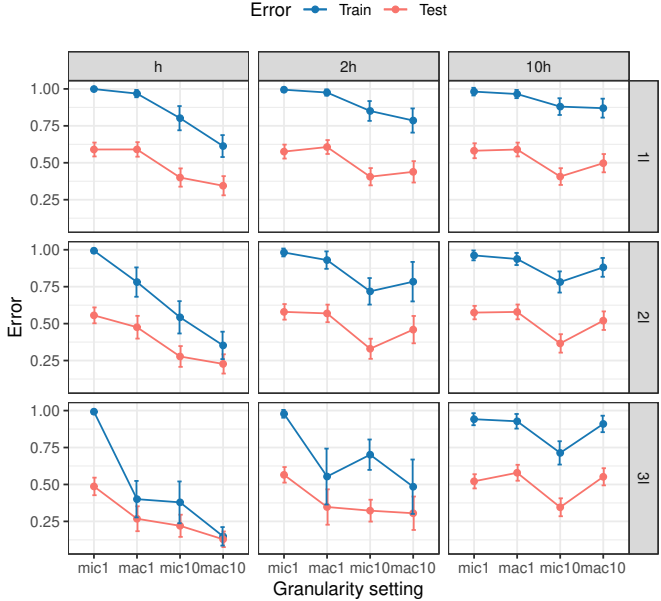}
	\caption{MNIST, classification accuracy for the training and test sets for the various NN architectures. Standard deviation for each point is reported as a vertical bar. Mic and mac correspond to micro and macro granularities, 1 and 10 correspond to $[-1,1]$ and $[-10,10]$ initialisation ranges. 1l, 2l and 3l correspond to 1, 2, and 3 hidden layers.}\label{fig:mnist:arch:class}
\end{figure}

The average classification accuracies arrived at by the progressive gradient walks are presented in Figure~\ref{fig:mnist:arch:class}. Averages were calculated across the accuracies as observed at the last step of each walk. The plots are grouped horizontally according to the layer width ($h, 2\times h$, and $10\times h$), and vertically according to the network depth (1, 2, and 3 hidden layers).  The results in Figure~\ref{fig:mnist:arch:class} show that the training accuracy ($C_t$) generally increased as the hidden layer width increased for all granularity settings except the $[-1,1]$ micro setting. Thus, the loss surfaces yielded by wider hidden layers were somewhat harder to exploit with very small steps, but the overall searchability, i.e., global landscape structure, improved. For the 1-hidden layer architecture, $C_t$ improved from $61\%$ to $87\%$ for the $[-10,10]$ macro setting as $h$ increased from $10$ to $100$. For the 3-hidden layer architecture, $C_t$ improved from $15\%$ to $91\%$ as $h$ increased from $10$ to $100$. This observation corresponds to a recent theoretical study by Johnson~\cite{ref:Johnson:2018}, where deep and ``skinny'' NN architectures, i.e., architectures with many hidden layers of a limited size, were shown to not be universal approximators. The generalisation accuracy ($C_g$) was also positively affected by an increase in width for the 2- and 3-hidden layer architectures.

\begin{figure*}[!htb]
	\begin{center}
		\begin{subfloat}[{$h = 10$, 1 hidden layer }]{    \includegraphics[width=0.31\textwidth]{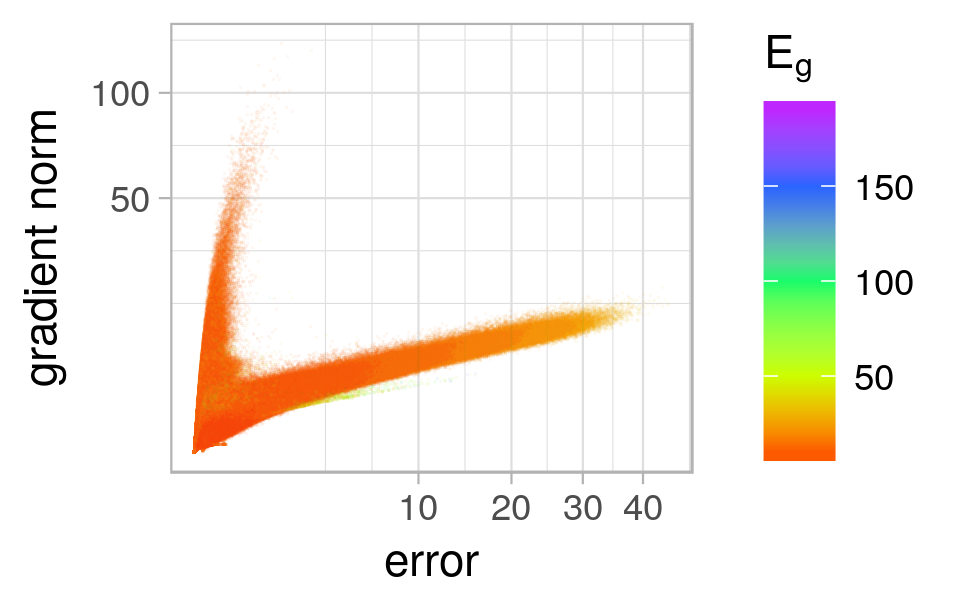}}
		\end{subfloat}
		\begin{subfloat}[{$2\times h = 20$, 1 hidden layer}]{    \includegraphics[width=0.31\textwidth]{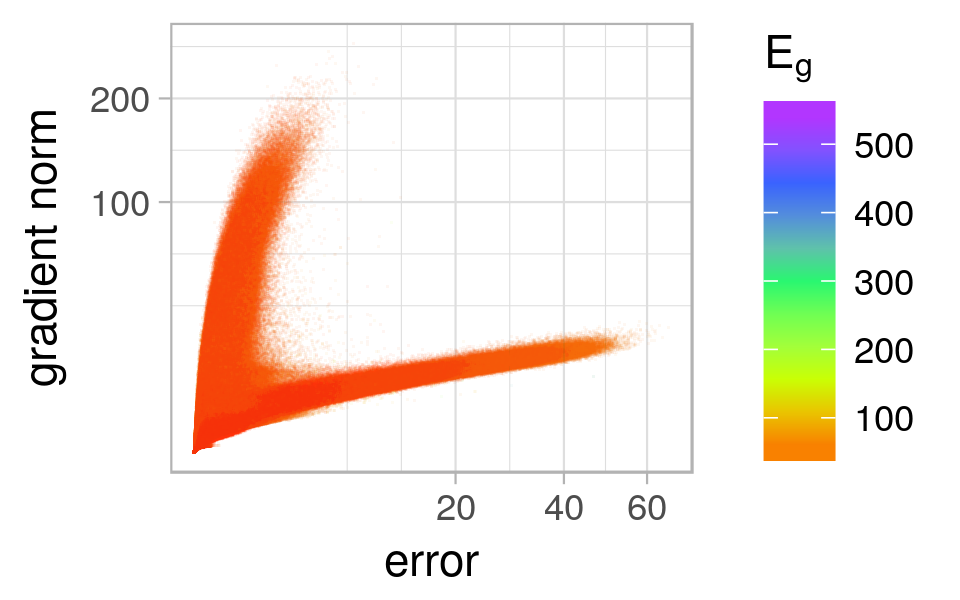}}
		\end{subfloat}
		\begin{subfloat}[{$10\times h = 100$, 1 hidden layer}]{    	\includegraphics[width=0.31\textwidth]{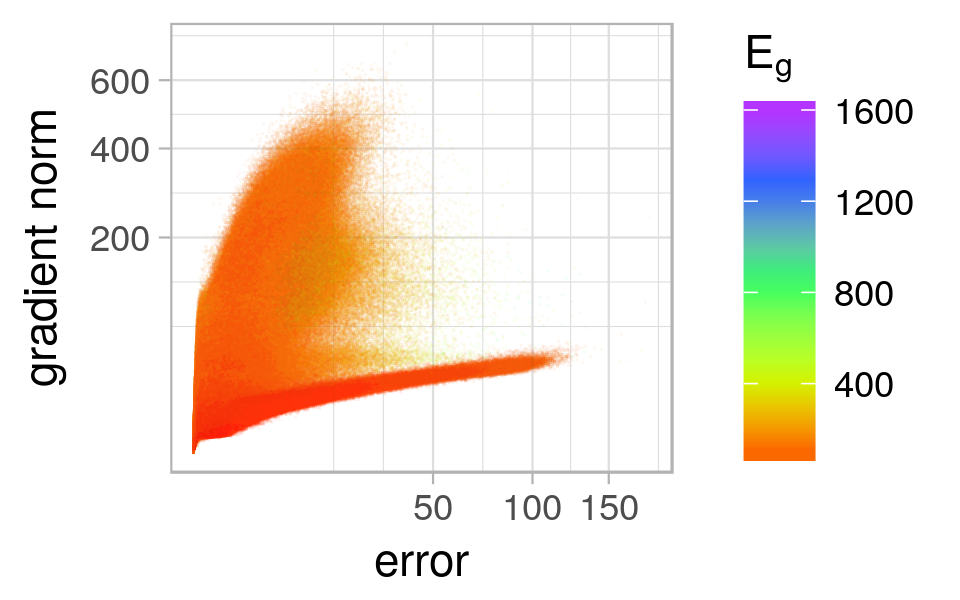}}
		\end{subfloat}	\\
		
		\begin{subfloat}[{$h = 10$, 2 hidden layers }]{    \includegraphics[width=0.31\textwidth]{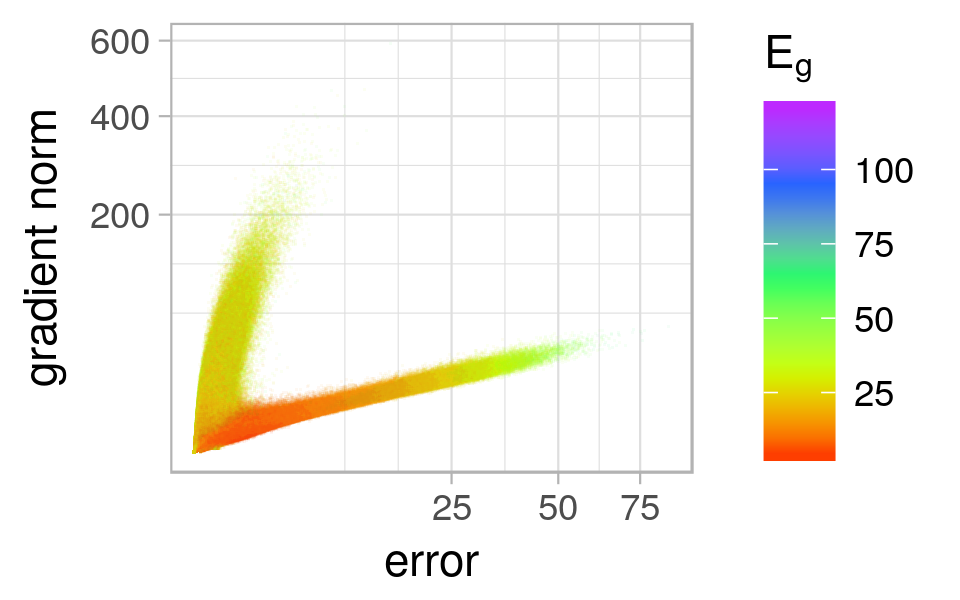}}
		\end{subfloat}
		\begin{subfloat}[{$2\times h = 20$, 2 hidden layers}]{    \includegraphics[width=0.31\textwidth]{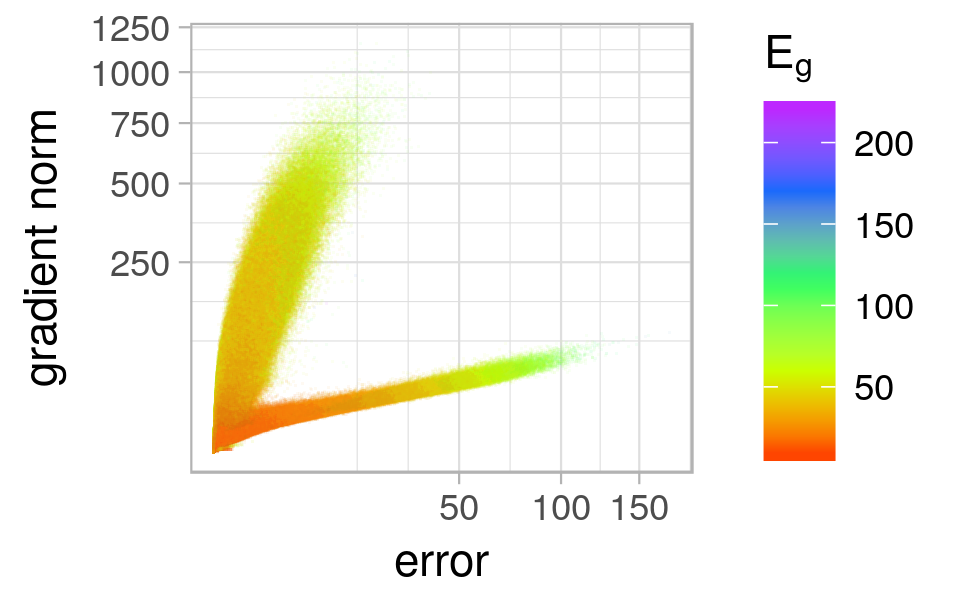}}
		\end{subfloat}
		\begin{subfloat}[{$10\times h = 100$, 2 hidden layers}]{    	\includegraphics[width=0.31\textwidth]{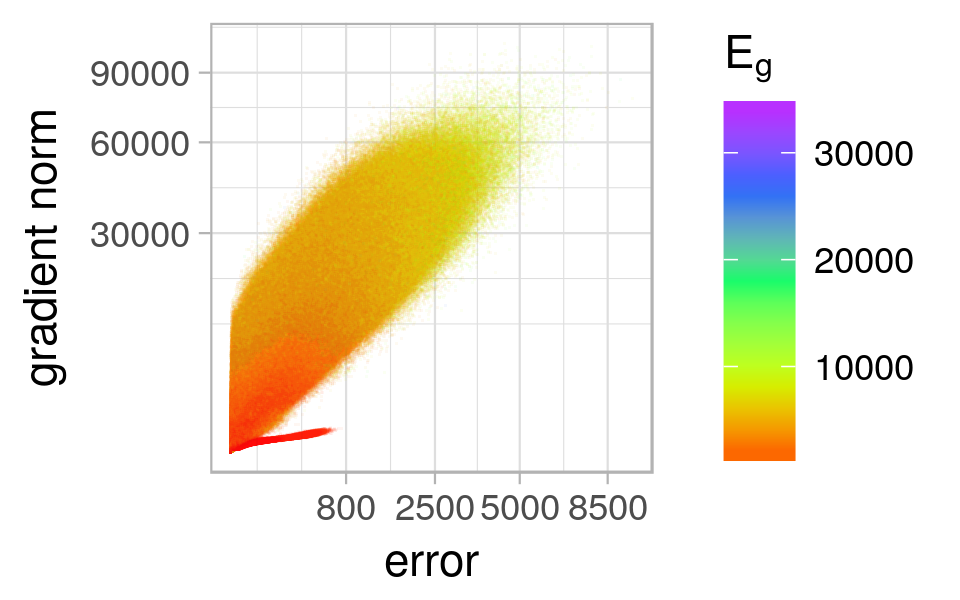}}
		\end{subfloat}\\
		
		\begin{subfloat}[{$h = 10$, 3 hidden layers  }]{    \includegraphics[width=0.31\textwidth]{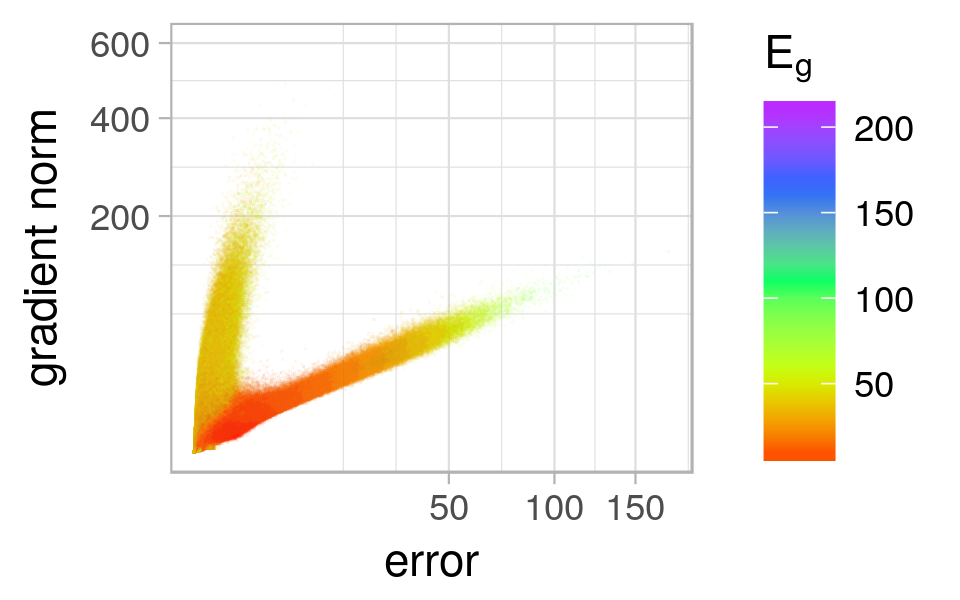}}
		\end{subfloat}	
		\begin{subfloat}[{$2\times h = 20$, 3 hidden layers}]{    \includegraphics[width=0.31\textwidth]{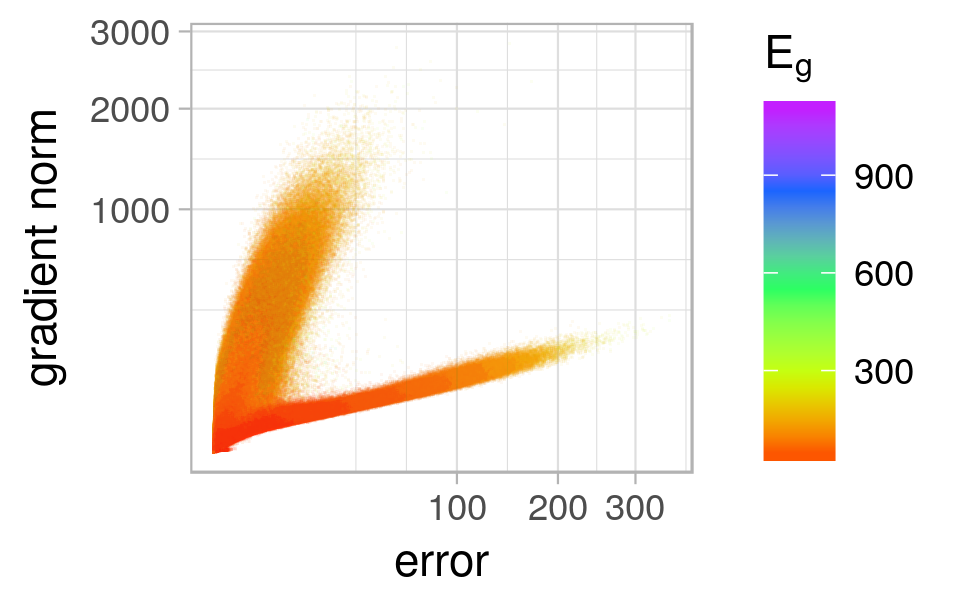}}
		\end{subfloat}	
		\begin{subfloat}[{$10\times h = 100$, 3 hidden layers}]{    \includegraphics[width=0.31\textwidth]{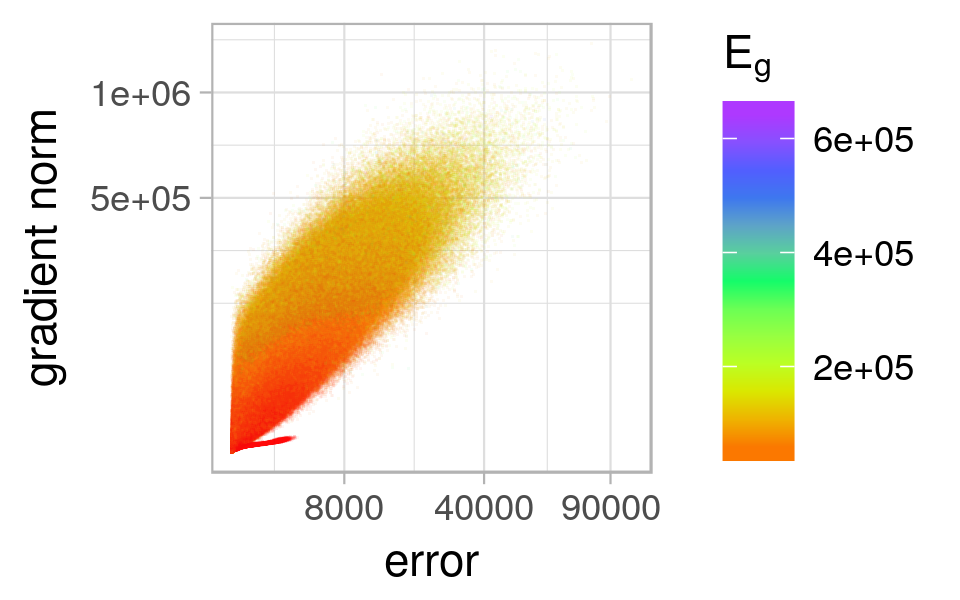}}
		\end{subfloat}
		\caption{Loss-gradient clouds, colourised according to the corresponding $E_g$ values, for the micro gradient walks ($\varepsilon = 0.02$) initialised in the $[-1,1]$ range for the MNIST problem.}\label{fig:mnist:arch:h:gen}
	\end{center}
\end{figure*}

Figure~\ref{fig:mnist:arch:h:gen} shows the l-g clouds obtained for the various architectures using the $[-1,1]$ micro walks, colourised according to the $E_g$ values. For all architectures considered, the sampled points split into the steep and shallow gradient clusters, where the steep gradient cluster generally corresponded to poor generalisation performance. These clusters are attributed to the narrow and wide valleys exhibited by the NN loss surfaces~\cite{ref:Chaudhari:2017,ref:Gallagher:2000, ref:Keskar:2017}. An increase in depth affected the shape of the steep gradient cluster by making it wider and steeper.  Figure~\ref{fig:mnist:arch:h:gen} shows that the transition from $h$ to $2\times h$, to $10\times h$, yielded the steep gradient cluster to become progressively wider. The overlap between the two clusters also increased with an increase in the architecture width.

\begin{figure*}[!htb]
	\begin{center}
		\begin{subfloat}[{$h = 10$, 1 hidden layer }]{    \includegraphics[width=0.31\textwidth]{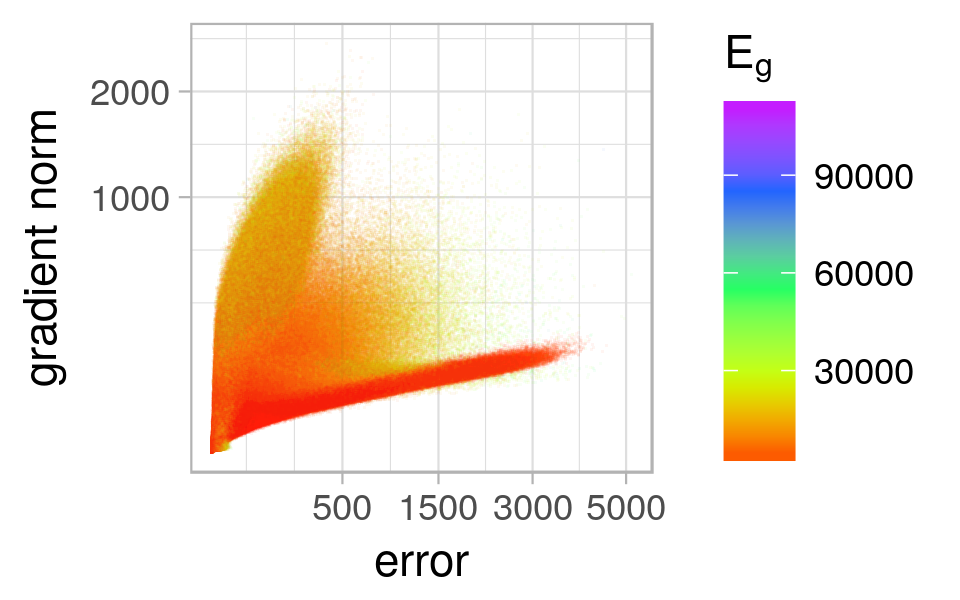}}
		\end{subfloat}
		\begin{subfloat}[{$2\times h = 20$, 1 hidden layer}]{    \includegraphics[width=0.31\textwidth]{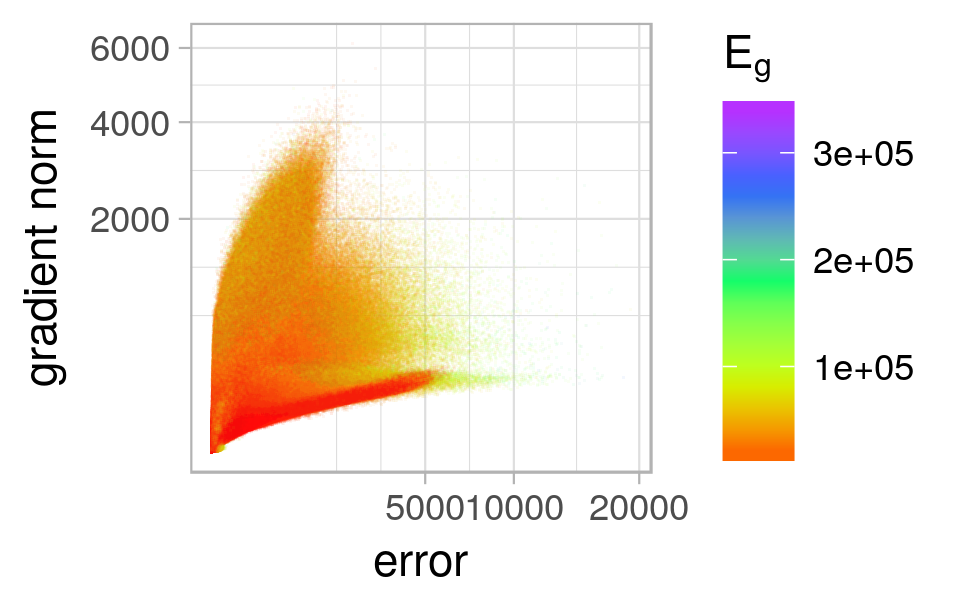}}
		\end{subfloat}
		\begin{subfloat}[{$10\times h = 100$, 1 hidden layer}]{    	\includegraphics[width=0.31\textwidth]{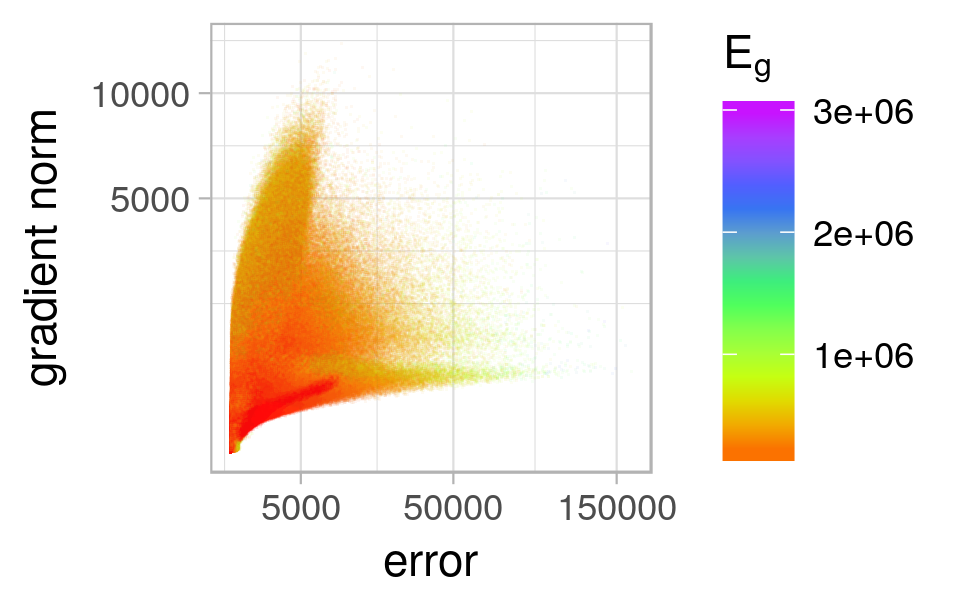}}
		\end{subfloat}	\\
		
		\begin{subfloat}[{$h = 10$, 2 hidden layers }]{    \includegraphics[width=0.31\textwidth]{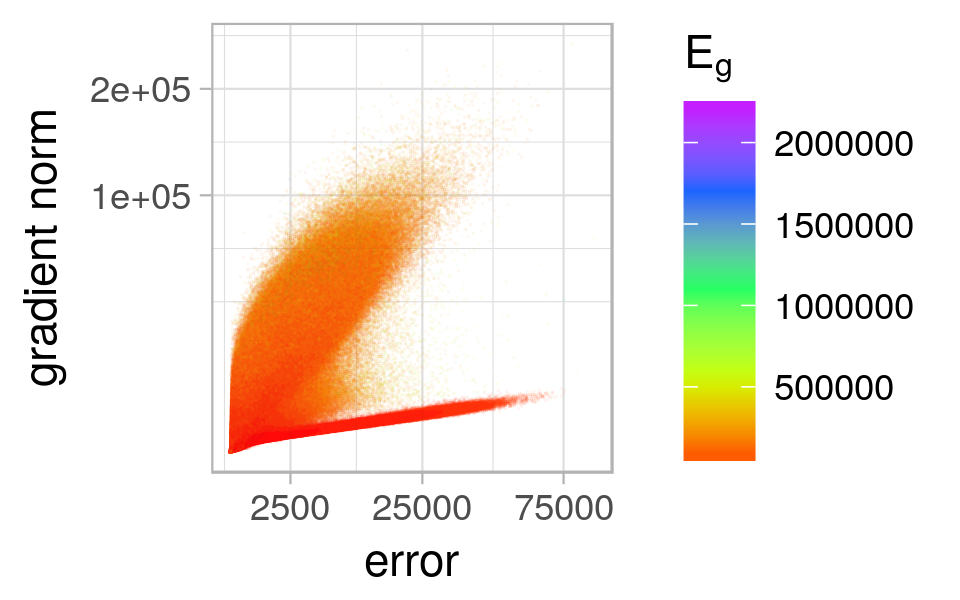}}
		\end{subfloat}
		\begin{subfloat}[{$2\times h = 20$, 2 hidden layers}]{    \includegraphics[width=0.31\textwidth]{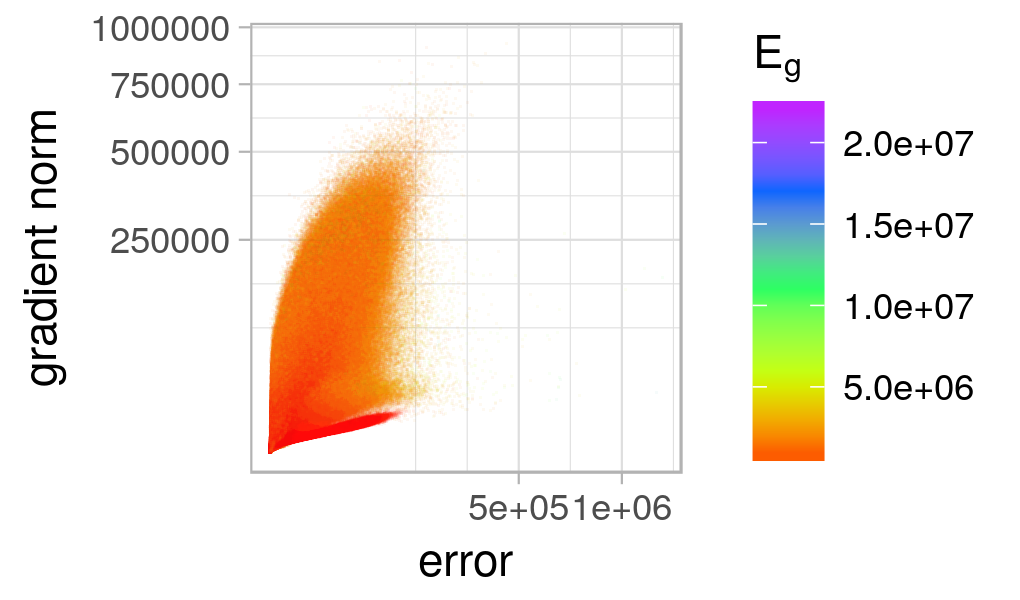}}
		\end{subfloat}
		\begin{subfloat}[{$10\times h = 100$, 2 hidden layers}]{    	\includegraphics[width=0.31\textwidth]{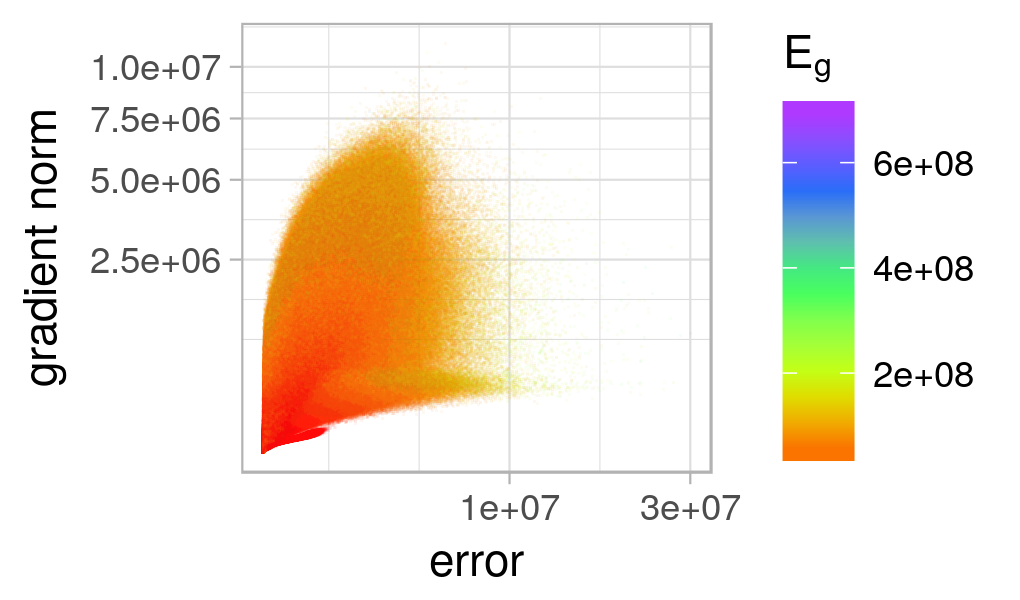}}
		\end{subfloat}\\
		
		\begin{subfloat}[{$h = 10$, 3 hidden layers  }]{    \includegraphics[width=0.31\textwidth]{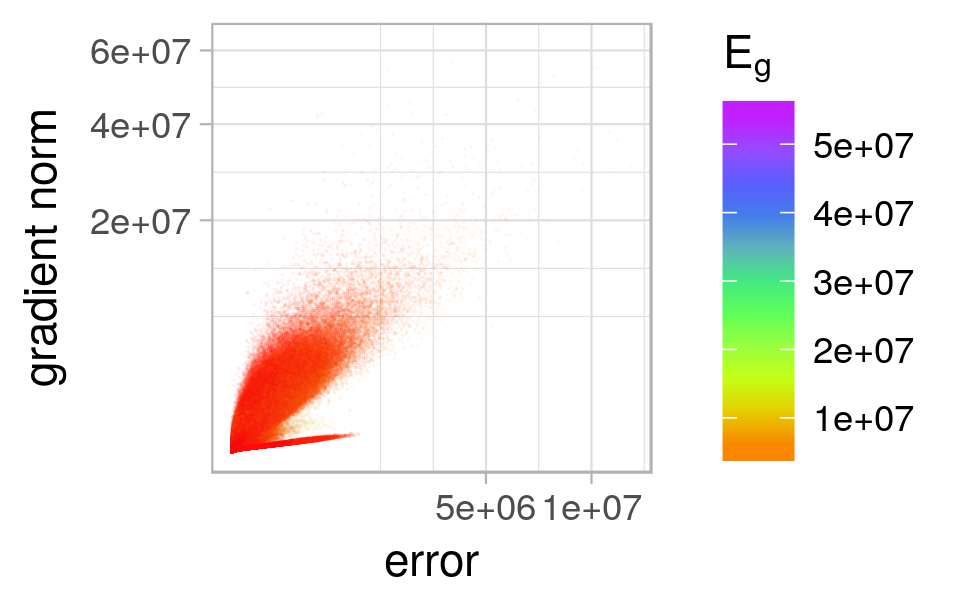}}
		\end{subfloat}	
		\begin{subfloat}[{$2\times h = 20$, 3 hidden layers}]{    \includegraphics[width=0.31\textwidth]{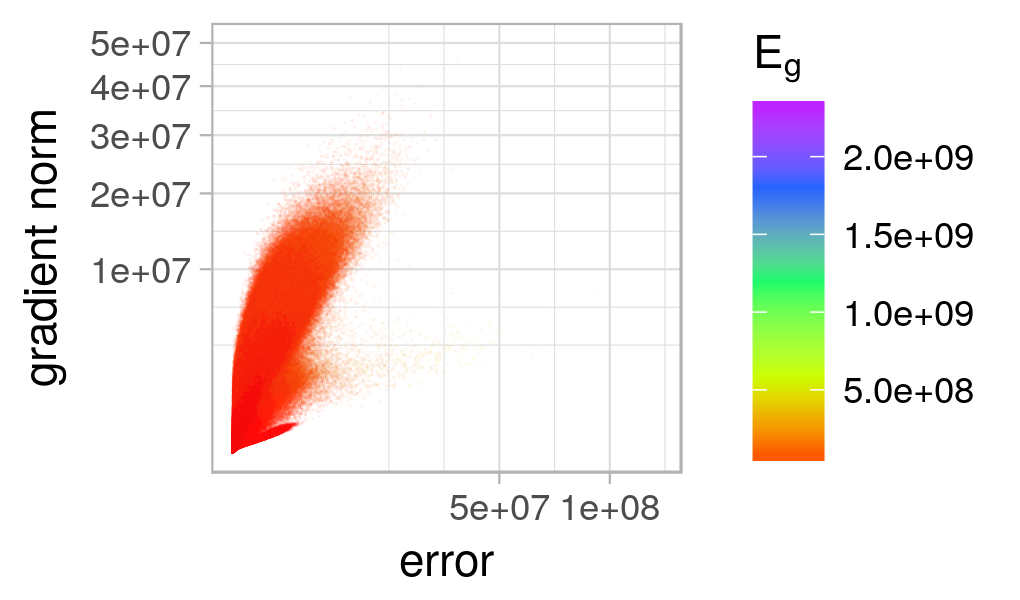}}
		\end{subfloat}	
		\begin{subfloat}[{$10\times h = 100$, 3 hidden layers}]{    \includegraphics[width=0.31\textwidth]{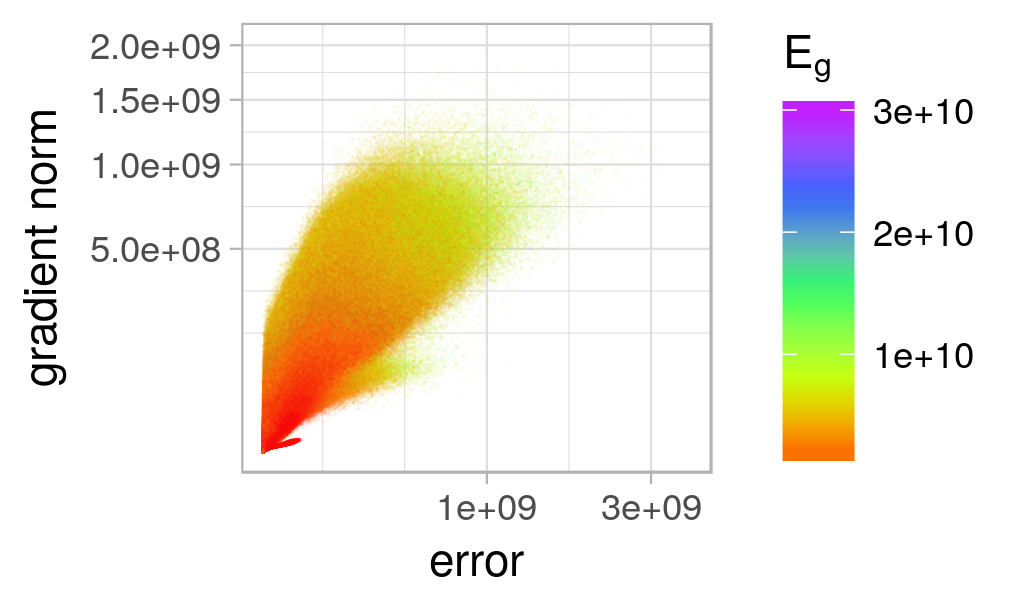}\label{fig:mnist:arch:h100:l3:b10:gen}}
		\end{subfloat}
		\caption{Loss-gradient clouds, colourised according to the corresponding $E_g$ values, for the micro gradient walks ($\varepsilon = 0.2$) initialised in the $[-10,10]$ range for the MNIST problem.}\label{fig:mnist:arch:h:b10:gen}
	\end{center}
\end{figure*}

Figure~\ref{fig:mnist:arch:h:b10:gen} shows the l-g clouds obtained by the $[-10,10]$ micro walks. For the wider initialisation range, the same split into two clusters is observed, although the two clusters appear more connected. The $[-10,10]$ micro walks used a larger maximum step size ($\varepsilon = 0.2$) than the $[-1,1]$ micro walks ($\varepsilon = 0.02$), which enabled the transition between the attractors. Overall, Figure~\ref{fig:mnist:arch:h:b10:gen} shows that an increase in width increased the overlap between the two clusters, effectively blending the two clusters into one wide cluster with a single global attractor at the error of zero. Thus, an increase in the hidden layer width simplified the loss surface.

Figure~\ref{fig:mnist:arch:class} shows that an increase in depth without an increase in width generally resulted in deteriorating $C_t$ and $C_g$ values. This observation once again correlates with the theoretical findings in~\cite{ref:Johnson:2018}. The l-g clouds in Figures~\ref{fig:mnist:arch:h:gen} and~\ref{fig:mnist:arch:h:b10:gen} show that an increase in depth increased the gradient and error magnitude ranges, but otherwise did not affect the degree of separation between the steep and the shallow gradient clusters. As the depth of the architecture increased, the steeper cluster became visibly heavier, and the shallow cluster diminished. Overall, an increase in depth did not simplify the structure of the MNIST loss surface, but made the global attractor steeper.

\section{Conclusions}\label{sec:arch:conclusions}
This paper presented a visual analysis of the NN loss surface modality associated with various NN architectures. Two different classification problems were considered. For each problem, $h$, $2\times h$, and $10 \times h$ hidden layer widths were considered, where $h$ corresponded to the minimal number of hidden neurons per layer. Further, for each hidden layer width, 1-, 2-, and 3-hidden layer architectures were considered. Each architecture was studied under four different granularity settings in order to capture the loss surface features present at different parts of the search space.

The results presented in this paper confirm that an increase in problem dimensionality yields an increase in indefinite, or flat curvature, as previously observed by Sagun et al.~\cite{ref:Sagun:2017}. An increase in NN depth yielded a more rapid increase in flatness than an increase in NN width. This behaviour is attributed to the inter-variable dependency between the hidden layers in a feed-forward architecture.

For the XOR problem, an increase in width was shown to reduce the number of local minima. For the same problem, an increase in depth was shown to reduce the convexity and increase the amount of saddle curvature. However, an increase in NN depth with a fixed width had no effect on the total number of stationary attractors. Thus, an increase in width was shown to change the shape of the attractor in a more meaningful way than an increase in depth. This observation correlates with~\cite{ref:Johnson:2018}, where deep and ``skinny'' NNs were shown to not exhibit the universal approximator properties.

For the MNIST problem, a single major attractor at the global minimum was observed. An increase in width, as well as an increase in depth, yielded an increase in the width and steepness of the observed attractor. A split into two clusters of steep and shallow gradients was observed. The clusters are attributed to the narrow and wide valleys exhibited by the NN loss surfaces~\cite{ref:Chaudhari:2017,ref:Gallagher:2000, ref:Keskar:2017}. An increase in width was shown to increase the overlap between the two clusters, up to a complete merge of the two clusters into a single cluster. An increase in depth did not exhibit the same effect. Instead, the steep cluster generally became heavier as more hidden layers were added, up to a complete disappearance of the shallow cluster. Thus, an increase in depth was shown to exaggerate the narrow valleys as a loss surface feature.

In general, both an increase in width, as well as an increase in depth, were shown to improve the searchability of the resulting loss surfaces. However, wider hidden layers were shown to be more instrumental in the overall improvement of the loss surface structure. Superior classification quality was associated with wider hidden layers. The steep gradient cluster was associated with inferior generalisation performance for the MNIST problem. 

Thus, an increase in the problem dimensionality was shown to yield a more searchable and more exploitable loss surface. An increase in width was shown to effectively reduce the number of local minima, and simplify the shape of the global attractor. An increase in depth was shown to sharpen the attractor, thus making it more exploitable.

In future, l-g clouds can be used to further improve our understanding of NN loss surfaces under various hyperparameter settings. In particular, the properties of wide and narrow valleys, and their relation to the global attractor, can be studied.

\section*{Acknowledgment}
The authors would like to thank the Centre for High Performance Computing (CHPC) (http://www.chpc.ac.za) for the use of their cluster to obtain the data for this study. 


\end{document}